%% file: main.tex
\documentclass[runningheads]{llncs}
\usepackage[T1]{fontenc}
\usepackage{amssymb}% http://ctan.org/pkg/amssymb
\usepackage{graphicx}
\usepackage{booktabs}
\usepackage{cleveref}
\usepackage{placeins}
\begin{document}
\title{ORacle: Large Vision-Language Models for Knowledge-Guided Holistic OR Domain Modeling}

\titlerunning{ORacle}
\author{Ege Özsoy\thanks{Equal contribution.}\inst{1,2}, Chantal Pellegrini\inst{\star,1,2}, Matthias Keicher\inst{1}, Nassir Navab\inst{1}}

\institute{
Computer Aided Medical Procedures, Technische Universit{\"a}t M{\"u}nchen, Germany\\
MCML, Germany
}
%

\maketitle              % typeset the header of the contribution
\begin{abstract}
Every day, countless surgeries are performed worldwide, each within the distinct settings of operating rooms (ORs) that vary not only in their setups but also in the personnel, tools, and equipment used. This inherent diversity poses a substantial challenge for achieving a holistic understanding of the OR, as it requires models to generalize beyond their initial training datasets. To reduce this gap, we introduce ORacle, an advanced vision-language model designed for holistic OR domain modeling, which incorporates multi-view and temporal capabilities and can leverage external knowledge during inference, enabling it to adapt to previously unseen surgical scenarios. This capability is further enhanced by our novel data augmentation framework, which significantly diversifies the training dataset, ensuring ORacle's proficiency in applying the provided knowledge effectively. In rigorous testing, in scene graph generation, and downstream tasks on the 4D-OR dataset, ORacle not only demonstrates state-of-the-art performance but does so requiring less data than existing models. Furthermore, its adaptability is displayed through its ability to interpret unseen views, actions, and appearances of tools and equipment. This demonstrates ORacle's potential to significantly enhance the scalability and affordability of OR domain modeling and opens a pathway for future advancements in surgical data science. We will release our code and data upon acceptance.

\end{abstract}

\input{chapters/introduction}

\input{chapters/method}
\input{chapters/experiments}

\input{chapters/conclusion}

\newpage

\bibliographystyle{splncs04}
\bibliography{refs}

\end{document}

%% file: chapters/introduction.tex
\section{Introduction}
\begin{figure}
\includegraphics[width=\textwidth]{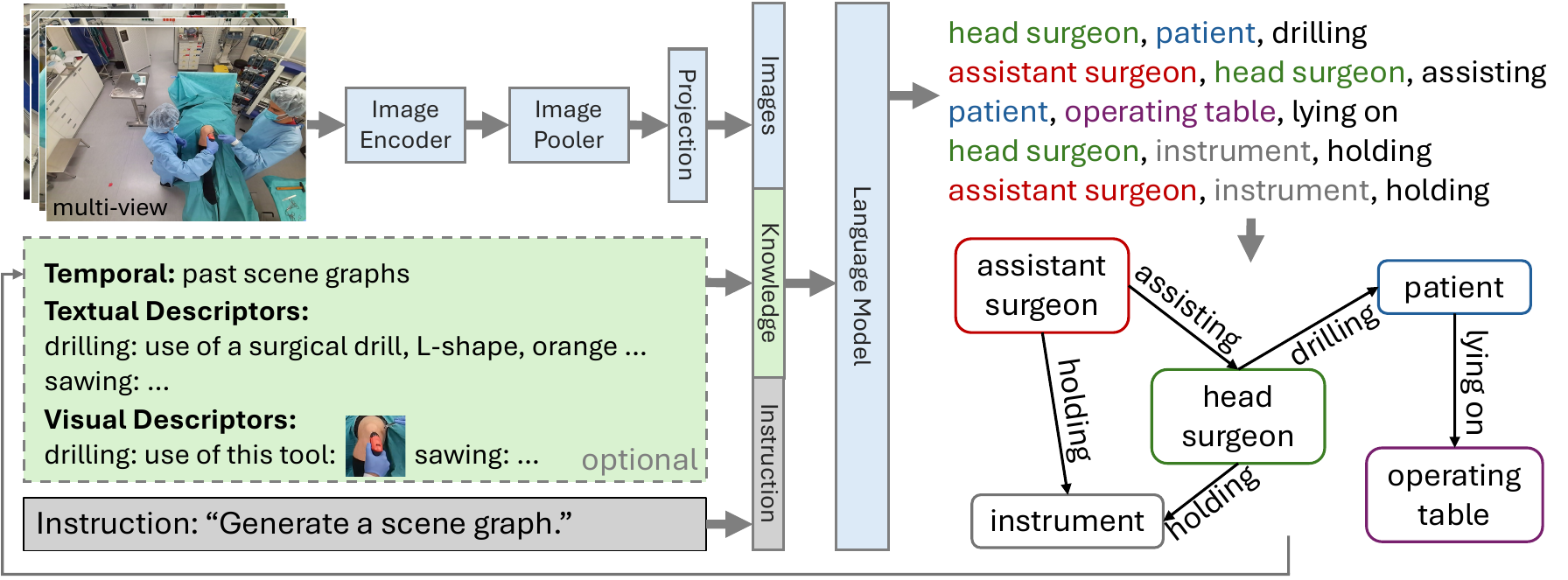}
\caption{An overview of our end-to-end scene graph generation architecture. ORacle takes as input multiview images and optionally additional knowledge and directly generates a scene graph token by token, considering all information at once.} \label{fig:model}
\end{figure}

The Operating Room (OR) is a dynamic and complex environment, demanding precise coordination, rapid decision-making, and flawless execution for successful surgical outcomes. While there is a critical need for automatic holistic modeling of the OR to enhance surgical data science (SDS)~\cite{lalys2014,maier-hein_sds_2017,ozsoy2023holistic}, the inherent variability and complexity of surgeries present significant challenges. These range from differences in workflows and setups to the diverse tools and staff involved, complicating the development of scalable models. In the field of SDS, many works have focused on specific tasks such as detecting the surgery phase~\cite{twinanda2016endonet,tecno}, actions~\cite{rendezvous,hao2023act,murali2023encoding,sharma2023surgical}, tools~\cite{jha2021kvasir,paranjape2023cross,wang2023pov}, identifying anatomies~\cite{das2023multi}, or visual question answering~\cite{surgicalgpt,bai2023revisiting}. Recently, Özsoy et al.~\cite{ozsoy20224d}, introduced the task of holistic OR domain modeling using semantic scene graphs, an accompanying dataset, 4D-OR, as well as a baseline method, which used the 3D point clouds and images from a timepoint to predict the scene graphs. In their follow-up work~\cite{ozsoy2023}, the authors proposed to use memory scene graphs for efficient modeling and integration of temporality and improved the accuracy and consistency of their results. However, these approaches face limitations in real-world applicability. Their reliance on expansive hardware setups, including six 3D calibrated RGB-D cameras, poses a barrier to widespread adoption, particularly in resource-limited settings. Moreover, these methods struggle with generalization, often failing to adapt to minor changes in the OR environment, such as variations in tool color.

Simultaneously, the advent of Large Vision-Language Models (LVLMs) has revolutionized the field of computer vision by demonstrating remarkable capabilities in understanding complex visual content~\cite{alayrac2022flamingo,dai2023instructblip,liu2023visualllava,liu2023improvedllava}. Their proficiency in processing and responding to natural language makes them ideally suited for integrating additional contextual knowledge. Yet, their potential in OR modeling remains untapped, partly due to the unique challenges of translating their capabilities to the nuanced and variable domain of surgical procedures.

To address the limitations in generalization and dependency on extensive hardware, we propose ORacle, a novel approach building upon the strengths of LVLMs for scalable and adaptable OR modeling. ORacle uniquely generates semantic scene graphs in an end-to-end manner directly from only multiview RGB images, circumventing the need for intermediate predictions or annotations such as human poses or object locations. To use multiple camera views efficiently and robustly, we introduce a multiview image pooler, which encodes a variable number of images from different perspectives into a unified representation. Furthermore, ORacle is the first method that supports the integration of multimodal knowledge, such as temporal information and detailed descriptors of OR tools and equipment, allowing adaptation to changes not seen during training without the need for extensive re-recording and retraining. To encourage an effective use of knowledge, we design an automatic data augmentation pipeline, enhancing the variability of 4D-OR. Lastly, we create and release a challenging digitally altered subset of the publicly available 4D-OR, enabling the evaluation of the adaptability and robustness of OR modeling approaches. ORacle achieves state-of-the-art results on scene graph generation on the 4D-OR dataset and sets a strong baseline for adaptability across varied settings. By eliminating the dependency on costly depth sensors, showcasing robust performance from even a single camera perspective, and enabling knowledge guidance, ORacle paves the way for accessible, cost-effective, and adaptable holistic OR modeling, with the potential to significantly impact surgical data science.\looseness=-1

%% file: chapters/method.tex
\section{Method}

\subsection{Scene Graph Generation as Language Modeling}
Scene graphs consist of nodes $N$, which correspond to entities in the scene, and edges $E$, which represent their relationships. In this work, we represent them as a list of triplets in the form of 
\textit{\textless subject, object, predicate\textgreater} where $subject, object \in N$ and $predicate \in E$. This list of triplets can be concatenated into one string, allowing us to model the scene graph generation as an image-to-sequence generation problem. 

\subsection{Architecture Overview}
The two main components of ORacle are a visual processing module and a large language model (LLM). The visual processing module converts images from one timepoint $T$ to a set of token embeddings in the feature space of the LLM. The LLM processes both the visual information and optionally additional knowledge, such as temporal history or textual and visual descriptors, and outputs a scene graph. A visualization of our architecture can be seen in Fig.~\ref{fig:model}.

\noindent \textbf{Visual Processing} We propose a novel, transformer-based~\cite{vaswani2017attention} image pooler to enable ORacle to process a variable number of images ${x_1,..., x_N}$ from multiple views. First, each image is encoded separately using a CLIP vision encoder~\cite{radford2021learning}. Then, the individual patch embeddings of all images are concatenated and the resulting sequence is processed by a transformer. Finally, we use the first N tokens as a joint representation, where N is fixed to the tokens representing one image. We then use an MLP to project this representation to the latent space of the LLM in the form of image tokens.

\noindent \textbf{Large Language Model} The image tokens are concatenated with the optional knowledge input and a scene graph generation instruction to form the input prompt. The LLM then autoregressively outputs a scene graph token by token.

During training, we randomly sample and shuffle the input views to ensure higher robustness to their number and order. Unlike previous methods~\cite{ozsoy20224d,ozsoy2023}, ORacle is trained entirely end-to-end, and learns to directly predict a scene graph from the input, without the need for object detection or pose prediction. 

\begin{figure}[tb]
\includegraphics[width=0.95\textwidth]{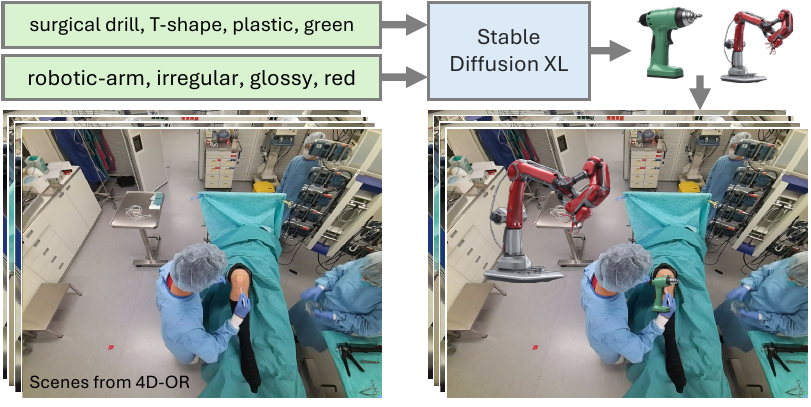}
\caption{An overview of our automatic variability enhancement pipeline, used during training. It first samples a set of attributes, then generates a matching object. Thereafter, it samples a suiting scene from 4D-OR and correctly places the sampled objects into it. For examples of more realistic scenes used during evaluation, see \Cref{fig:qual_results}.} \label{fig:data_generation}
\end{figure}

\subsection{Knowledge Integration}
We design ORacle to allow the integration of knowledge, such as temporal context or tool and equipment descriptors, improving performance and adaptability.

\noindent \textbf{Temporal Context}
To efficiently integrate the surgery's history, we condense the scene graphs from previous time points into a surgery change log, represented as a sequence of triplets. Inspired by LABRAD-OR~\cite{ozsoy2023}, we handle the long and short-term history differently. For the long-term history, we include every triplet only once in the order they are first predicted, providing an approximate understanding of where the current time point is in the surgery. For the short-term history, we include all predictions of the last five time points, letting the model additionally know if actions are repeated and for how long they are performed. This temporal knowledge is included as text in the prompt.

\noindent \textbf{Textual Descriptors}
The goal of textual descriptors is to inform the network of appearances of tools and equipment by describing their attributes in text form. We define an attribute space consisting of the following five attributes: \texttt{\{object type, color, size, shape, texture\}}. In this space, we can flexibly describe relations and entities. Such descriptors allow to inform the network about new or different-looking objects during inference time without the need for retraining.

\noindent \textbf{Visual Descriptors}
With textual descriptors, it can be difficult to accurately describe appearances. Therefore, we additionally introduce the concept of visual descriptors, where we include a single image of the respective tool or equipment in the prompt, encoded using the frozen CLIP vision encoder. The global CLIP embedding is projected into the LLM space, such that one object is represented as a single token in the language space.

\noindent \textbf{Symbolic Scene Graph Representation}
We want to enable our model to work with unseen entities and predicates during inference in an open vocabulary setting and pay more attention to the descriptors. To accomplish this, in our adaptable models, we represent all objects and relations in a symbolic space. For instance instead of \textit{\textless head surgeon, patient, drilling\textgreater}, we represent a triplet as \textit{\textless $SUB$, $OBJ$, $PRED$\textgreater} with $ SUB, OBJ \in \{A, B,... , Z\}$ and $ PRED \in \{\alpha, \beta, ...\omega\}$. For every sample, we randomly pick symbols for all objects and predicates and pair every symbol with its descriptor in the prompt. This forces the model to rely on the descriptors instead of class names to make a correct prediction. At inference time, we can guide the model to understand appearance changes and recognize novel objects or predicates by adapting the descriptors.

\noindent \textbf{Automatic Variability Enhancement} The limited variability of tools and equipment in 4D-OR hinders effective teaching of knowledge descriptor use. To address this, we developed an automatic variability enhancement pipeline, depicted in~\Cref{fig:data_generation}. First, we sample attributes describing a surgical tool or equipment. Using stable diffusion XL~\cite{podell2023sdxl}, we generate an image that corresponds to these attributes and remove its background using DIS~\cite{qin2022}. Then, we select a time point from 4D-OR and place the synthetic object in all views of the scene. For placing tools, we leverage the hand location of surgeons and place them on top. Equipment is placed over large objects in the scene, such as the anesthesia machine or tables. In total, we use 200,000 such samples for training. Although the resulting placements do not always appear as realistic as the scenes we are evaluating on, we show that this still significantly enhances the model's usage of the accompanying textual and visual descriptors within the knowledge prompt.

%% file: chapters/experiments.tex
\begin{table}[b]
    \caption{Distribution of included entities and relations in our adaptability benchmark. Relations and entities not existing in 4D-OR are marked with *.}
  \centering
    \begin{tabular}{c|c|c|c|c|c|c|c|c|c}
    \toprule
    &  {cementing} & {cutting} & {drilling} & {hammering} & {sawing} & {suturing} & {Kuka*} & {Mako*} & {rob. sawing*}\\\midrule
    Nr. & 12 & 12 & 12 & 12 & 12 & 12 & 10 & 10 & 10  \\
    \bottomrule
  \end{tabular}
  \label{tab:ad_benchmark}
\end{table}
\section{Experiments}
We conduct experiments to evaluate our models' scene graph generation performance and adaptability to unseen objects, predicates, appearances, and views.

\noindent \textbf{Dataset}
We train and evaluate on 4D-OR~\cite{ozsoy20224d} with the official data splits. 4D-OR includes ten takes of simulated partial knee replacement surgeries, recorded from 6 views with RGB-D cameras. ORacle only uses four views without depth. All 6734 scenes are labeled with semantic scene graphs, phases, and human roles.

\noindent \textbf{Adaptability Benchmark} To objectively evaluate adaptability, we created a set of 102 manually digitally altered 4D-OR images. This benchmark ensures photo-realism by accounting for the appropriate context, orientation, and occlusions, distinguishing it from the outputs of our automatic pipeline. It comprises images captured from two distinct views and features a spectrum of tools and equipment ranging from familiar yet visually altered to entirely novel types. Each sample consists of a positive image with the digital alteration and a negative image without the alteration, as well as a prompt textually and visually describing the adaptations. The included relations and entities can be found in \Cref{tab:ad_benchmark}.

\noindent \textbf{Evaluation Metrics}
Following previous works, for scene graph prediction, we evaluate our model using macro F1 over all relation types and temporal consistency between the time points. Additionally, we report precision and recall for evaluation on our adaptability benchmark.

\noindent \textbf{Implementation Details}
We initialize our LVLM with the weights of LLaVA-7B~\cite{liu2023visualllava}, with a CLIP~\cite{radford2021learning} based image encoder and Vicuna-7B~\cite{vicuna2023} as the LLM. We fully finetune the image feature extractor and finetune the LLM using LoRA~\cite{hu2021lora} on a Nvidia A-40 GPU. We train all our models until convergence, either on the original 4D-OR dataset or the samples generated by our automatic variability enhancement pipeline. For the temporal models, we closely follow LABRAD-OR~\cite{ozsoy2023}, including using curriculum learning, whereby a model previously trained without temporal information is further finetuned using temporal knowledge.

\begin{table}[tb]
\begin{minipage}{.50\textwidth}
    \centering
    \caption{Scene graph generation results on the 4D-OR test set. SV and MV indicate single view or multiview input, respectively.}
    \begin{tabular}{@{}l|c|c|c|c@{}}
    \toprule
                     & MV & Depth & Temporal  & F1  \\ \midrule
    4D-OR & $\checkmark$ & $\checkmark$ & $\times$ &  0.83         \\ 
    LABRAD-OR & $\checkmark$ & $\checkmark$ & $\checkmark$ & 0.88         \\  \midrule
    ORacle-SV & $\times$  & $\times$ & $\times$ & 0.84        \\ 
    ORacle-SV-T & $\times$ & $\times$  & $\checkmark$ & 0.86         \\ 
    ORacle-MV & $\checkmark$ & $\times$ & $\times$ & 0.88         \\  
    ORacle-MV-T & $\checkmark$ & $\times$ & $\checkmark$ & \textbf{0.91}         \\ \bottomrule 
    \end{tabular}
    \label{tab_main_results}
\end{minipage}\hspace*{0.7cm}
\begin{minipage}[]{.40\textwidth}
    \centering
    \caption{Ablation study on adaptable models, showing using symbolic representation and adaptable knowledge representations do not lead to worse results on the 4D-OR test set.}
    \begin{tabular}{@{}l|c@{}}
    \toprule
     &  F1  \\ \midrule 
    ORacle-MV & 0.88         \\  
    ORacle-adapt-Text & 0.87         \\ 
    ORacle-adapt-Vis & 0.88         \\  \bottomrule 
    \end{tabular}
    \label{tab_adapt_results}
\end{minipage}
\end{table}

\begin{table}[b]
\begin{minipage}{.4\textwidth}
    \centering
    \caption{Results on our adaptability benchmark.}
    \begin{tabular}{@{}l|c|c|c@{}}
    \toprule
                     & Prec & Rec & F1  \\ \midrule
    ORacle-MV & 0.86 & 0.22 & 0.31         \\ 
    ORacle-adapt-Text & 0.83  & 0.78 & 0.78        \\ 
    ORacle-adapt-Vis & 0.92  & 0.63 & 0.71       \\ \bottomrule
    \end{tabular}
    \label{tab:adapt_benchmark}
\end{minipage}\hspace*{0.7cm}
\begin{minipage}[]{.5\textwidth}
    \centering
    \caption{Results when only using view two (F1-2) or six (F1-6). "noAug" indicates training without order augmentation.}
    \begin{tabular}{@{}l|c|c@{}}
    \toprule
                    & F1-2  & F1-6   \\ \midrule
    4D-OR & 0.50 & 0.20       \\ 
    ORacle-MV-noAug & 0.84 & 0.47 \\
    ORacle-MV & 0.87 & 0.62       \\  \bottomrule
    
    \end{tabular}
    \label{tab:view_robustness}
\end{minipage}
\end{table}

\begin{figure}
\includegraphics[width=0.95\textwidth]{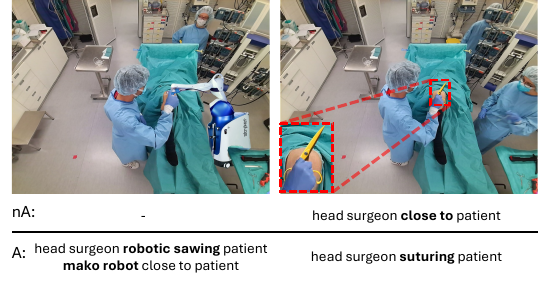}
\caption{Results on our adaptability benchmark of the non-adaptable ORacle-MV model (nA) and our adaptable models (A). Left: ORacle-adapt-text; Right: ORacle-adapt-vis.} \label{fig:qual_results}
\end{figure}

\section{Results and Discussion}

\noindent \textbf{Scene Graph Generation}
We compare ORacle to previous works on 4D-OR in \Cref{tab_main_results}. Our best-performing model, relying on four RGB images and temporal context, outperforms the previous SOTA, which additionally relies on two more RGB images and six depth images, by 3\%. ORacles' predictions also are temporally consistent, reaching a consistency score of 0.89 compared to 0.87 in LABRAD-OR~\cite{ozsoy2023}, getting very close to the ground truth consistency of 0.9. Without temporality, we outperform 4D-OR~\cite{ozsoy20224d} by 5\%, and get comparable results even when relying on one single view. Overall, the results show that both temporality and multi-view integration positively affect performance. In \Cref{tab_adapt_results}, we show the results of our adaptable model variants, demonstrating that performance remains similar when using the symbolic models with descriptors.

\noindent \textbf{Downstream Tasks}
We evaluate our model on clinical role prediction and surgical phase recognition. Unlike prior works, ORacle always outputs a scene graph already including clinical roles, alleviating the need to explicitly solve clinical role prediction. It achieves a macro F1 of 0.85, which is only slightly worse than 0.89 from LABRAD-OR~\cite{ozsoy2023}, which relies on ad-hoc analyses, including tracking and a secondary role prediction model. We also compare our model to the existing SOTA in surgical phase recognition, closely following their methodology~\cite{ozsoy2023holistic}. Without adapting any of the heuristics for mapping scene graph sequences to surgical phases, we achieve an F1 score of 0.99, compared to their result of 0.97.

\noindent \textbf{Adaptability}
We validate the robustness and adaptability of our method when using textual and visual descriptors, depicted as ORacle-adapt-text and ORacle-adapt-vis, respectively. We work in a single time point setting, therefore comparing the non-temporal ORacle models to the 4D-OR method~\cite{ozsoy20224d}.

\noindent \textit{Single View:} \Cref{tab:view_robustness} analyzes the robustness to using a single view during inference. We evaluate all models on only view two, which was included in the training views, and see that ORacle shows high robustness to this single-view setting, reaching an F1 score of 0.87, which is only 1\% less than using all four views. Secondly, we compare performance on view six, which is a novel view for ORacle, not used during training. Even though 4D-OR did use this view in training, ORacle performs significantly better, demonstrating its high robustness to novel views. Further, we ablate the effect of order augmentations during multiview training, confirming that these are crucial for increasing robustness.

\noindent \textit{Novel appearances and objects:} \Cref{tab:adapt_benchmark} shows the results on our adaptability benchmark, and \Cref{fig:qual_results} shows qualitative examples. Both adaptable models perform significantly better than the non-adaptable model variant. This shows both robustness to appearance changes of known tools and equipment as well as to novel ones. We also digitally altered one take of 4D-OR, switching the color of the drill and the saw in all frames. While both 4D-OR and ORacle-MV get confused in this setting, our adaptable models, especially when using visual descriptors, are substantially more robust towards this change, as shown in \Cref{tab:color}. Lastly, we test the understanding of novel relations by training ORacle on a modified training set, excluding all time points in which the drilling or sawing relation occurs. Prior works and our non-adaptable model, cannot predict relations that were not seen during training. Our adaptable models show a good performance on these relations as shown in \Cref{tab:novel_attr}.

\begin{table}[t]
\begin{minipage}{.45\textwidth}
    \centering
    \caption{Results when the colors of the drill (orange) and saw (green) were digitally altered to be the exact opposite of each other only during evaluation. "no switching" is provided as a baseline value.} 
    \begin{tabular}{@{}l|c|c@{}}
    \toprule
                      & drill-F1 & saw-F1  \\ \midrule
    ORacle-MV no switching & 0.94 &  0.98\\ \midrule
    4D-OR & 0.06  & 0.06       \\ 
    ORacle-MV & 0.0  & 0.0       \\  
    ORacle-adapt-Text & 0.08  & 0.35      \\ 
    ORacle-adapt-Vis  & 0.74  & 0.79     \\ \bottomrule
    
    \end{tabular}
    \label{tab:color}
\end{minipage}\hspace*{0.7cm}
\begin{minipage}[]{.45\textwidth}
    \centering
    \caption{Results when training without drilling or sawing scenes. "full training" corresponds to using these scenes for training. Both 4D-OR and ORacle-MV can not predict any unseen predicates, indicated by N/A.}  
    \begin{tabular}{@{}l|c|c@{}}
    \toprule
                      & drill-F1 & saw-F1  \\ \midrule
    ORacle-MV full training & 0.94 &  0.98\\ \midrule
    4D-OR & N/A  & N/A       \\ 
    ORacle-MV & N/A  & N/A       \\ 
    ORacle-adapt-Text & 0.44  & 0.49      \\ 
    ORacle-adapt-Vis  & 0.79  & 0.52     \\ \bottomrule
    
    \end{tabular}
    \label{tab:novel_attr}
\end{minipage}
\end{table}

%% file: chapters/conclusion.tex
\FloatBarrier
\section{Conclusion}
In this work, we propose ORacle, an adaptable knowledge-guided holistic OR modeling approach building upon Large Vision-Language Models. Our method generates semantic scene graphs end-to-end from multiview RGB images, by employing a novel multiview image pooler. We introduce multimodal knowledge guidance, which for the first time, allows adapting to previously unseen concepts in inference time. Our results show that ORacle reaches state-of-the-art results not only in the benchmark task of scene graph generation but also demonstrates significantly improved adaptability to unseen OR scenarios. By reducing the requirement for extensive data collection and annotations, we believe ORacle opens up a pathway for more affordable and scalable OR domain modeling.\looseness=-1

\section{Acknowledgements}
This work has been partially supported by Stryker, the EVUK programme ("Next-generation Al for Integrated Diagnostics”) of the Free State of Bavaria, Bundesministerium für Bildung und Forschung (BMBF) with grant [ZN 01IS17050].

%% file: main.bbl
\begin{thebibliography}{10}
\providecommand{\url}[1]{\texttt{#1}}
\providecommand{\urlprefix}{URL }
\providecommand{\doi}[1]{https://doi.org/#1}

\bibitem{alayrac2022flamingo}
Alayrac, J.B., Donahue, J., Luc, P., Miech, A., Barr, I., Hasson, Y., Lenc, K., Mensch, A., Millican, K., Reynolds, M., et~al.: Flamingo: a visual language model for few-shot learning. Advances in Neural Information Processing Systems  \textbf{35},  23716--23736 (2022)

\bibitem{bai2023revisiting}
Bai, L., Islam, M., Ren, H.: Revisiting distillation for continual learning on visual question localized-answering in robotic surgery. In: International Conference on Medical Image Computing and Computer-Assisted Intervention. pp. 68--78. Springer (2023)

\bibitem{vicuna2023}
Chiang, W.L., Li, Z., Lin, Z., Sheng, Y., Wu, Z., Zhang, H., Zheng, L., Zhuang, S., Zhuang, Y., Gonzalez, J.E., Stoica, I., Xing, E.P.: Vicuna: An open-source chatbot impressing gpt-4 with 90\%* chatgpt quality (March 2023)

\bibitem{tecno}
Czempiel, T., Paschali, M., Keicher, M., Simson, W., Feussner, H., Kim, S.T., Navab, N.: Tecno: Surgical phase recognition with multi-stage temporal convolutional networks. In: Medical Image Computing and Computer Assisted Intervention - {MICCAI} 2020 - 23nd International Conference, Shenzhen, China, October 4-8, 2020, Proceedings, Part {III}. Lecture Notes in Computer Science, vol. 12263, pp. 343--352. Springer (2020)

\bibitem{dai2023instructblip}
Dai, W., Li, J., Li, D., Tiong, A.M.H., Zhao, J., Wang, W., Li, B., Fung, P., Hoi, S.: Instructblip: Towards general-purpose vision-language models with instruction tuning (2023)

\bibitem{das2023multi}
Das, A., Khan, D.Z., Williams, S.C., Hanrahan, J.G., Borg, A., Dorward, N.L., Bano, S., Marcus, H.J., Stoyanov, D.: A multi-task network for anatomy identification in endoscopic pituitary surgery. In: International Conference on Medical Image Computing and Computer-Assisted Intervention. pp. 472--482. Springer (2023)

\bibitem{hao2023act}
Hao, L., Hu, Y., Lin, W., Wang, Q., Li, H., Fu, H., Duan, J., Liu, J.: Act-net: Anchor-context action detection in surgery videos. In: International Conference on Medical Image Computing and Computer-Assisted Intervention. pp. 196--206. Springer (2023)

\bibitem{hu2021lora}
Hu, E.J., Shen, Y., Wallis, P., Allen-Zhu, Z., Li, Y., Wang, S., Wang, L., Chen, W.: Lora: Low-rank adaptation of large language models. arXiv preprint arXiv:2106.09685  (2021)

\bibitem{jha2021kvasir}
Jha, D., Ali, S., Emanuelsen, K., Hicks, S.A., Thambawita, V., Garcia-Ceja, E., Riegler, M.A., de~Lange, T., Schmidt, P.T., Johansen, H.D., et~al.: Kvasir-instrument: Diagnostic and therapeutic tool segmentation dataset in gastrointestinal endoscopy. In: MultiMedia Modeling: 27th International Conference, MMM 2021, Prague, Czech Republic, June 22--24, 2021, Proceedings, Part II 27. pp. 218--229. Springer (2021)

\bibitem{lalys2014}
Lalys, F., Jannin, P.: Surgical process modelling: a review. International Journal of Computer Assisted Radiology and Surgery, Springer Verlag  \textbf{9},  495--511 (2014)

\bibitem{liu2023improvedllava}
Liu, H., Li, C., Li, Y., Lee, Y.J.: Improved baselines with visual instruction tuning. arXiv preprint arXiv:2310.03744  (2023)

\bibitem{liu2023visualllava}
Liu, H., Li, C., Wu, Q., Lee, Y.J.: Visual instruction tuning. arXiv preprint arXiv:2304.08485  (2023)

\bibitem{maier-hein_sds_2017}
Maier-Hein, L., Vedula, S.S., Speidel, S., Navab, N., Kikinis, R., Park, A., Eisenmann, M., Feussner, H., Forestier, G., Giannarou, S., Hashizume, M., Katic, D., Kenngott, H., Kranzfelder, M., Malpani, A., März, K., Neumuth, T., Padoy, N., Pugh, C., Schoch, N., Stoyanov, D., Taylor, R., Wagner, M., Hager, G.D., Jannin, P.: Surgical data science for next-generation interventions. Nature Biomedical Engineering  \textbf{1}(9),  691–696 (Sep 2017)

\bibitem{murali2023encoding}
Murali, A., Alapatt, D., Mascagni, P., Vardazaryan, A., Garcia, A., Okamoto, N., Mutter, D., Padoy, N.: Encoding surgical videos as latent spatiotemporal graphs for object and anatomy-driven reasoning. In: International Conference on Medical Image Computing and Computer-Assisted Intervention. pp. 647--657. Springer (2023)

\bibitem{rendezvous}
Nwoye, C.I., Yu, T., Gonzalez, C., Seeliger, B., Mascagni, P., Mutter, D., Marescaux, J., Padoy, N.: Rendezvous: Attention mechanisms for the recognition of surgical action triplets in endoscopic videos. Medical Image Analysis  \textbf{78} (2022)

\bibitem{ozsoy2023}
{\"O}zsoy, E., Czempiel, T., Holm, F., Pellegrini, C., Navab, N.: Labrad-or: Lightweight memory scene graphs for accurate bimodal reasoning in dynamic operating rooms. In: International Conference on Medical Image Computing and Computer-Assisted Intervention. Springer (2023)

\bibitem{ozsoy2023holistic}
{\"O}zsoy, E., Czempiel, T., {\"O}rnek, E.P., Eck, U., Tombari, F., Navab, N.: Holistic or domain modeling: a semantic scene graph approach. International Journal of Computer Assisted Radiology and Surgery  (2023). \doi{10.1007/s11548-023-03022-w}, \url{https://doi.org/10.1007/s11548-023-03022-w}

\bibitem{ozsoy20224d}
{\"O}zsoy, E., {\"O}rnek, E.P., Eck, U., Czempiel, T., Tombari, F., Navab, N.: 4d-or: Semantic scene graphs for or domain modeling. In: Medical Image Computing and Computer Assisted Intervention--MICCAI 2022: 25th International Conference, Singapore, September 18--22, 2022, Proceedings, Part VII. Springer (2022)

\bibitem{paranjape2023cross}
Paranjape, J.N., Sikder, S., Patel, V.M., Vedula, S.S.: Cross-dataset adaptation for instrument classification in cataract surgery videos. In: International Conference on Medical Image Computing and Computer-Assisted Intervention. pp. 739--748. Springer (2023)

\bibitem{podell2023sdxl}
Podell, D., English, Z., Lacey, K., Blattmann, A., Dockhorn, T., M{\"u}ller, J., Penna, J., Rombach, R.: Sdxl: Improving latent diffusion models for high-resolution image synthesis. arXiv preprint arXiv:2307.01952  (2023)

\bibitem{qin2022}
Qin, X., Dai, H., Hu, X., Fan, D.P., Shao, L., Gool, L.V.: Highly accurate dichotomous image segmentation. In: ECCV (2022)

\bibitem{radford2021learning}
Radford, A., Kim, J.W., Hallacy, C., Ramesh, A., Goh, G., Agarwal, S., Sastry, G., Askell, A., Mishkin, P., Clark, J., et~al.: Learning transferable visual models from natural language supervision. In: International conference on machine learning. pp. 8748--8763. PMLR (2021)

\bibitem{surgicalgpt}
Seenivasan, L., Islam, M., Kannan, G., Ren, H.: Surgicalgpt: End-to-end language-vision gpt for visual question answering in surgery. In: International Conference on Medical Image Computing and Computer-Assisted Intervention. pp. 281--290. Springer (2023)

\bibitem{sharma2023surgical}
Sharma, S., Nwoye, C.I., Mutter, D., Padoy, N.: Surgical action triplet detection by mixed supervised learning of instrument-tissue interactions. In: International Conference on Medical Image Computing and Computer-Assisted Intervention. pp. 505--514. Springer (2023)

\bibitem{twinanda2016endonet}
Twinanda, A.P., Shehata, S., Mutter, D., Marescaux, J., De~Mathelin, M., Padoy, N.: Endonet: a deep architecture for recognition tasks on laparoscopic videos. IEEE transactions on medical imaging  \textbf{36}(1),  86--97 (2016)

\bibitem{vaswani2017attention}
Vaswani, A., Shazeer, N., Parmar, N., Uszkoreit, J., Jones, L., Gomez, A.N., Kaiser, {\L}., Polosukhin, I.: Attention is all you need. Advances in neural information processing systems  \textbf{30} (2017)

\bibitem{wang2023pov}
Wang, R., Ktistakis, S., Zhang, S., Meboldt, M., Lohmeyer, Q.: Pov-surgery: A dataset for egocentric hand and tool pose estimation during surgical activities. In: International Conference on Medical Image Computing and Computer-Assisted Intervention. pp. 440--450. Springer (2023)

\end{thebibliography}
